\documentclass[10pt,twocolumn,letterpaper]{article}

\pdfoutput=1
\usepackage{wacv}
\usepackage{times}
\usepackage{epsfig}
\usepackage{graphicx}
\usepackage{amsmath}
\usepackage{amssymb}
\usepackage{booktabs}
\usepackage{xcolor}
\usepackage{graphicx}
\usepackage{multirow}
\usepackage{subcaption}
\usepackage[accsupp]{axessibility}

\newcommand{\burst}{\{I_i\}}
\newcommand{\x}{\mathbf{x}}
\newcommand{\valid}{\{V_i\}}

\newcommand{\bgframes}{\{B_i\}}
\newcommand{\fgframes}{\{F_i\}}

\newcommand{\flowsref}{\{f_{kj}\}}
\newcommand{\segmentations}{\{S_i\}}
\newcommand{\keyframe}{\tilde B_k}
\newcommand{\sota}{state-of-the-art}
\newcommand{\fim}{RDN}
\newcommand{\spynet}{SPyNet}
\newcommand{\soldpt}{SOLD$_{pt}$}
\newcommand{\soldtf}{SOLD$_{tf}$}
\newcommand{\pwcnet}{PWC-Net}
\newcommand{\unet}{U-net}
\newcommand{\fps}{7~fps}

\newcommand{\res}{$320\times192$}

\wacvalgorithmstrack   % Uncomment this line if you are submitting to the Algorithms Track.
\wacvfinalcopy % *** Uncomment this line for the final submission

\usepackage[pagebackref=true,breaklinks=true,colorlinks,bookmarks=false]{hyperref}
\begin{document}

%%%%%%%%% TITLE
\title{Efficient Flow-Guided Multi-frame De-fencing}

\author{Stavros Tsogkas \and Fengjia Zhang \and Allan Jepson   \and Alex Levinshtein
\and
Samsung AI Center Toronto\\
101 College St., Toronto, ON, Canada, M5G 1L7\\
{\tt\small \{stavros.t, f.zhang2, allan.jepson, alex.lev\}@samsung.com}
}

\maketitle
\thispagestyle{empty}

\begin{abstract}
Taking photographs ``in-the-wild'' is often hindered by fence obstructions that stand between the camera user 
and the scene of interest, and which are hard or impossible to avoid. 
De-fencing is the algorithmic process of automatically removing such obstructions from images, revealing 
the invisible parts of the scene.
While this problem can be formulated as a combination of fence segmentation and image inpainting,
this often leads to implausible hallucinations of the occluded regions. 
Existing multi-frame approaches rely on propagating information to a selected keyframe from its temporal neighbors, 
but they are often inefficient and struggle with alignment of severely obstructed images.
In this work we draw inspiration from the video completion literature, and develop a simplified 
framework for multi-frame de-fencing that computes high quality flow maps directly from obstructed frames,
and uses them to accurately align frames.  
Our primary focus is efficiency and practicality in a real world setting: 
the input to our algorithm is a  short image burst (5 frames) --
a data modality commonly available in modern smartphones-- and the output is a 
single reconstructed keyframe, with the fence removed. 
Our approach leverages simple yet effective CNN modules, trained on carefully generated synthetic data, 
and outperforms more complicated alternatives real bursts, both quantitatively and qualitatively, 
while running real-time.

\end{abstract}

\section{Introduction} \label{sec:intro}

\begin{figure}[t]
    \centering
    \includegraphics[width=\linewidth]{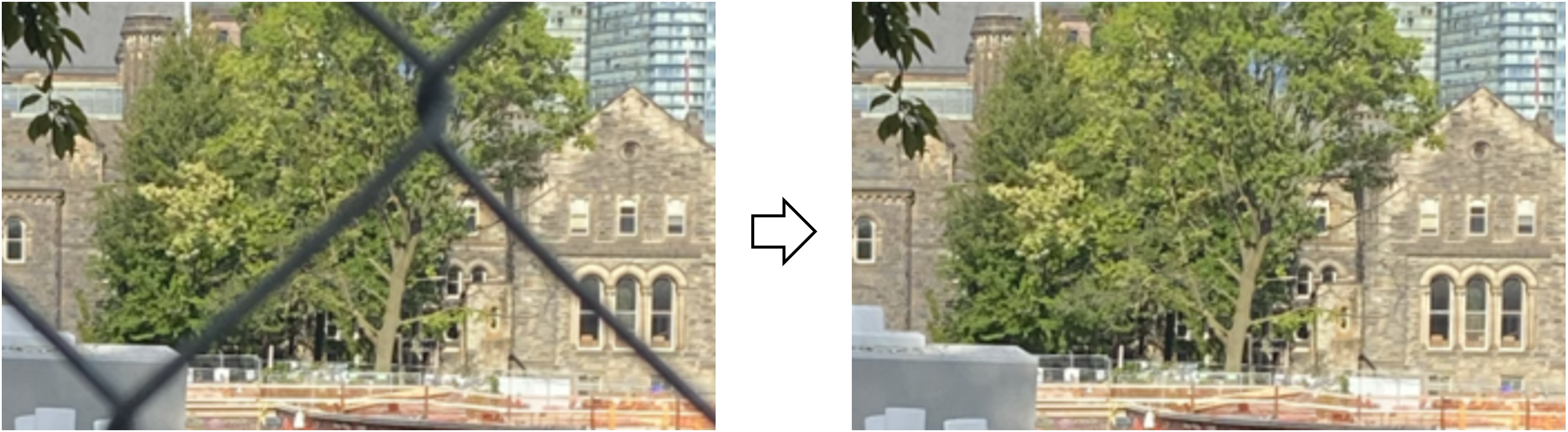}
    \caption{We train a simple and efficient model for \textbf{de-fencing}: removing fence obstructions from images, revealing the underlying scene of interest.
    Our method is fast and can accurately remove fences of varying size and appearance from \emph{real} bursts, without online finetuning.}
    \label{fig:teaser}
\end{figure}
Rapid improvements in both the camera hardware and image processing software have turned modern cell phones into powerful yet portable image and video recording devices.
% \st{add a couple of citations on statistics regarding cell phone use as image/video recording tools?}
% Modern cell phones are powerful computing machines that allow users to run demanding applications, consume media content, and take high-quality photographs and videos.
% The latter, in particular, is enabled by rapid improvements in both the camera hardware and image processing algorithms, which have dramatically improved image quality, while keeping the size of these devices manageable.
This has enabled and encouraged 
casual users to shoot photos without any time for special preparation, setup, or framing of the shot.
On the flip side, photos and videos taken under these conditions rarely contain just the object(s) of interest,
and are hindered by various obstructions that stand between the subject and the user.
% All these are examples of \emph{obstructions} that alter the quality of the scene we are 
% trying to memorize\st{apathanatisoume} and that are either very difficult or completely impossible
% to avoid.

One type of obstruction that is of special interest because of its commonness is \emph{fences}.
Imagine, for instance, taking a photo of an animal through a zoo fence, or people playing basketball in a fenced-out outdoor court;
these are just a few everyday scenes that are obstructed by fence structures that are either inconvenient or completely impossible to avoid.
\emph{De-fencing} uses computer vision algorithms to automatically remove such fence obstructions from images, as shown in Figure~\ref{fig:teaser}.
De-fencing is a harder problem than it may initially seem.
Fences have varying structure and appearance patterns, and come in different thicknesses and sizes. 
Furthermore, reconstruction of the background scene can become challenging because of low lighting and noise, or motion blur, caused by rapidly moving objects. 

The first works that tackled de-fencing in a principled manner were by Liu et al.~\cite{liu2008image,park2010image}. 
They formulate the problem as the segmentation of a repeating foreground pattern that exhibits 
approximate translational symmetry, followed by inpainting to recover the occluded image regions.
The approach of \cite{liu2008image} is mainly limited by the fact that it uses a single image as input.
Because of the opaque nature of the fence obstruction, the occluded parts of the scene must be hallucinated by the inpainting algorithm;
\cite{park2010image} partially addresses this by using a photo taken from a different view, to reduce the number of pixels that must be hallucinated.

De-fencing can also be viewed as a special case of the more general problem of layer separation~\cite{pawan2008learning, alayrac2019visual, gandelsman2019double, liu2021learning},
which models an image as a composition of individual layers, e.g., a foreground 
layer containing the obstruction and a background layer containing the scene of interest.
Xue et al.~\cite{xue2015computational} formulate generic obstruction removal as a layer separation problem, driven by motion parallax.
Although their solution is generic and works well, it involves multiple time-consuming, hand-tuned optimization steps,
and the use of hand-crafted motion and image priors. 
SOLD~\cite{liu2021learning} is a deep learning re-incarnation of~\cite{xue2015computational} 
that achieved \sota{} results on obstruction removal, and can also be adapted to remove fences from 
a multi-frame input burst.
Unfortunately, SOLD depends on computationally expensive networks for flow computation and frame reconstruction,
making it impractical for use in low-powered devices.
It also often requires an online optimization step that takes $\sim3$ minutes to produce acceptable results on real bursts, 
and even without this input-specific finetuning, it cannot be run in real time.
Finally, since the background frames are the output of a reconstruction CNN module, they 
occasionally contain inconsistencies or artifacts. 
% \aj{online (?) although see comment here $\rightarrow$} 
% AJ: I don't understand the point of this sentence if you really mean learning-based approaches are impractical for phones, which is what is said. (BTW the current context is SOLD, not general learning based methods.) Did you mean approaches that require on-line optimization? You also say "sizable" CNN's.. not sure what that means; "sizable" with respect to what?  In general, can you make this sentence more concrete, and tie it to SOLD? Something like SOLD doesn't fit into current cell phones, and requires on-line opt. Or we demonstrate a smaller and faster model that outperforms it both in terms of memory and runtime requirements.

Flow-guided completion methods~\cite{huang2016temporally, xu2019deep, xu2019deep, li2022towards, zhang2022inertia} 
reduce such artifacts by computing inpainted flow maps between pairs of obstructed frames and using them to  explicitly transfer pixel values to a reference 
frame from its temporal neighbors.
Inpainting flows in the occluded area is easier  than directly inpainting pixel values, 
so a less powerful network can be used, and the results tend to look more plausible because 
the pixel values are taken from real frames instead of being hallucinated by a generative network.
These works do not make any assumptions regarding the shape and type of the occlusion, other the fact that it is completely opaque.
However, the mask marking the occluded area is considered to be known, which is an unrealistic requirement for our purposes.

Here our goal is to develop a de-fencing algorithm that prioritizes \emph{efficiency} and \emph{practicality}. 
We develop a framework that enjoys the realism and modularity of flow-based video completion approaches, 
while being significantly simpler to train and deploy.
Instead of videos, the input to our algorithm is shorter bursts of $K=5$ frames, 
which are a very common photo modality in modern smartphones.
The type of occlusion is known (fences), but we do not make any assumptions regarding its spatial extent or location;
instead, we train a class-specific segmentation model to automatically detect fences in images.
Computing flow maps of scenes occluded by fences, presents us with a 
new challenge, as standard optical flow networks fail under the presence of repeated patterns~\cite{jonna2016deep,liu2021learning}.
To solve this problem, we train a segmentation-aware \spynet{}~\cite{ranjan2017optical} that can simultaneously compute and inpaint
flow maps corresponding to the occluded background scene, ignoring foreground occlusions.
Finally, to quantitatively evaluate the performance of our approach on \emph{real data}, 
we collect a dataset of multi-frame sequences and corresponding ``pseudo-groundtruth'' for the reference frame using a alignment procedure, akin to~\cite{bhat2021deep}.
To summarize:
\begin{itemize}
    \item We design a CNN pipeline for multi-frame de-fencing that is simple, modular, efficient, and easy to train.
    % consists of a small number of easily trainable independent components.\aj{or easily trainable? see comment above}
    \item Unlike flow-based works, which assume the occlusion is known, we estimate it automatically from the input.
    \item We train a segmentation-aware optical flow model that can reliably estimate flows corresponding to the background scene despite severe fence obstructions.
    \item Our method achieves \sota{} results on synthetic and real bursts, without requiring sequence-specific finetuning. As a result, it has significantly lower runtimes 
    compared to alternatives.
\end{itemize}

\section{Related Work} \label{sec:related}

\subsection{Image and Video De-fencing} \label{sec:related:defencing}
Liu et al.~\cite{liu2008image} is perhaps the first work to formally introduce de-fencing in a computer vision context, as symmetry-driven automatic fence segmentation, followed by inpainting.
% \aj{We could remove the next two sentences, by adding "with automatic fence segmentation followed by inpainting" to the end of the previous sentence.}
% They start by discovering a lattice that captures the approximate 
% translational symmetry of the fence in an single input image.
% Pixels are classified as foreground or background using appearance cues and background holes are filled with texture inpainting.
\cite{park2010image} improves on this work by using online learning to aid lattice detection and segmentation, 
and by leveraging a second viewpoint to improve inpainting.
Jonna et al.~\cite{jonna2015multimodal,jonna2017stereo} also improve fence segmentation, by complementing RGB with depth data.

\cite{mu2013video, yi2016automatic} extend de-fencing to video sequences of arbitrary number of frames.
Mu et al~\cite{mu2013video} rely on motion parallax to separate foreground fence obstructions from background (although the 
definition of ``fence'' is quite loose), 
while Yi et al.~\cite{yi2016automatic} describe a bottom-up approach for video de-fencing that 
groups pixels in each frame using color and motion cues. 
Both methods rely on optimization techniques to refine their optical flow or frame inpainting results.

% They also estimate fence masks and motion fields used to perform sub-pixel frame alignment, but their 
% pipeline is composed of a series of optimization stages, and involves selection of quite a few hyperparameters. \aj{Are we arguing that our approach should be preferred, at least in part, because Mu et al uses "quite a few" hyperparameters? :)}
% Yi et al.~\cite{yi2016automatic} describe a bottom-up approach for video de-fencing that 
% groups pixels in each frame using color and motion cues. 
% Solving a graph-cut optimization produces an initial fence labelling that is later refined using a CRF~\cite{krahenbuhl2011efficient}
% and existing hole-filling algorithms are used to complete the segmented area. 
% \aj{Maybe the previous two methods can be more briefly discussed as methods that require on-line optimization, and joined into one initial para of this section.}

More recently, deep learning has been adopted for video defencing~\cite{jonna2016deep,du2018accurate}.
Jonna et al.~\cite{jonna2016deep} use a pretrained classification CNN as a feature extractor and train
a SVM classifier that distinguishes fence from non-fence patches.
The authors reformulate an existing optical flow algorithm to make it  occlusion-aware and recover the
de-fenced image using FISTA optimization~\cite{beck2009fast}. 
Du et al.~\cite{du2018accurate} replace the CNN-SVM combination 
with a fully convolutional network (FCN)~\cite{long2015fully} and apply 
temporal refinement to the extracted segmentations by aggregating information from neighboring frames.
Our approach shares a similar pipeline but simplifies both the segmentation extraction and occlusion-aware
flow computation steps, while being considerably faster, as we do not perform test-time optimization.

\subsection{Layer Separation} \label{sec:related:layer_separation}
A more generic formulation of the fence removal problem  
views an image as a composition of layers, each with its own alpha map (which can be semi transparent), and the goal is to separate the layers. 
In~\cite{gandelsman2019double} the foreground-background layers are the output of two convolutional network, 
trained per image in an unsupervised fashion, 
and are recovered using a deep image prior~\cite{ulyanov2018deep}.
A similar idea is used by Alayrac et al.~\cite{alayrac2019visual} for video decomposition, but with supervised training. 
Other approaches for video decomposition use explicit motion information~\cite{xue2015computational,liu2021learning}.
Xue et al.~\cite{xue2015computational} describe a computational method for decomposing a scene
into an foreground obstruction layer and a background scene from multi-scale motion cues of a multi-frame sequence, in an unsupervised way.
They solve an optimization problem that alternatingly finds the constituent layers and the respective
motion fields that, when used to align the burst to a reference frame, can reconstruct the original frames with low error.
A modern reincarnation of this approach is proposed by Liu et al. with SOLD~\cite{liu2021learning}.
SOLD follows a similar multi-scale approach, using a convolutional framework, both for layer reconstruction
and motion estimation -- the latter wth a pre-trained PWC-Net~\cite{sun2018pwc}.

% \st{
% - alayrac2019visual \al{done}
% - nandoriya2017video
% - pawan2008learning
% }

\subsection{Flow-based Video Completion} \label{sec:related:video_completion}
Video completion is a related problem to multi-frame fence removal, 
the main difference being that the segmentation mask is assumed to be provided and the emphasis is typically on longer frame sequences. 
Our work is motivated by Xu et al.~\cite{xu2019deep}, who proposed the idea of first tackling the easier problem of flow inpainting,
and then using the completed flows to propagate color values to a reference frame from its temporal neighbors.
Since it not guaranteed that all occluded pixels are visible in some frame, a separate
\emph{image} inpainting step must be used to fill any remaining holes.
Gao et al.~\cite{gao2020flow} improve this approach by synthesizing sharp flow edges along the 
object boundaries and using non-local temporal neighborhoods for propagating pixels across frames. 
These works involve a series of individual, separately trained processing stages, some of which are hand-crafted, inefficient, 
and can potentially compromise performance of subsequent stages. 
Li et al.~\cite{li2022towards} address this issue by proposing an end-to-end framework for flow-guided video completion.
Their approach was developed concurrently to our own and shares some of its simplicity and efficiency advantages. 
However, their framework still operates under the assumption that the occlusion mask is provided.
% and our approach has the potential to be more efficient for de-fencing 
% \st{must read their timings more carefully and polish this. also, should we keep this here or move to the related work section?}

% \st{
% - huang2016temporally
% - zhang2022inertia
% }

% Perhaps the most closely related works are by Du et al.~\cite{du2018accurate} and \cite{li2022towards}. 
% \cite{du2018accurate} has a similar pipeline to ours, conceptually.
% However, we did not find temporal refinement of the fence segmentations necessary, and we replace their reconstruction stage 
% with a more efficient and effective learned convolutional module. 
% \cite{li2022towards} was concurrent work with ours, and addresses some of the shortcomings of previous flow-based methods, such 
% as replacing multiple individually trained stages with an end-to-end framework.

\section{Method} \label{sec:method}
\begin{figure*}[t]
    \centering
    \includegraphics[width=\textwidth]{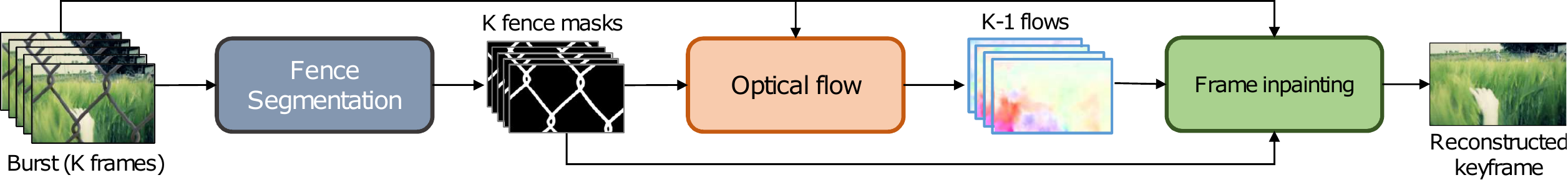}
    \caption{Given a burst of $K$ frames with fence obstructions as input, we reconstruct 
        a single keyframe, after removing the fence.
        Our pipeline is composed of three distinct steps: 
        a) initially, a fence mask is estimated on each input frame individually with a \unet{} fence segmentation model (Sec.~\ref{sec:method:fence_seg});
        b) the estimated masks are used to condition a segmentation-aware optical flow \spynet{}$^m$, which simultaneously computes and inpaints flows corresponding only to the background scene, ignoring the repeated fence occlusion patterns (Sec.~\ref{sec:method:flow_mask_conditional});
        c) finally, an image inpainting module takes the estimated masks and flows, aligns the frames with respect to a selected keyframe and fills in the missing pixel values (Sec.~\ref{sec:method:frame_inpainting}).
    }
    \label{fig:pipeline}
\end{figure*}

We begin with an overview of the problem we are solving, and establish the notation that we  use throughout the paper.
The input to our algorithm is a burst of $K$ RGB frames, $\burst{}$, composed from an unknown background scene of interest$\bgframes{}$ 
and an unknown opaque foreground occlusion in the form of a fence $\fgframes{}$. Specifically,
\begin{equation}
    I_i = S_i \cdot F_i + (1 - S_i) \cdot B_i,
    \label{eq:layer_composition}
\end{equation}

\noindent
where $S_i \in [0, 1]$ is a soft fence occlusion mask. Our goal is to train a model that removes the fence obstruction from $\burst{}$ 
and recovers a single keyframe background image $B_{k}$, where $k$ is the keyframe index.
%reveals the occluded parts of the scene \emph{for a specific keyframe} in the sequence.

Instead of outputting the unobstructed frame directly, we break down the problem into the steps illustrated in Figure~\ref{fig:pipeline}.
We start by training a network that is applied individually on each frame in $\burst{}$, and outputs fence segmentation predictions $\segmentations{}$.
The role of these segmentations is two-fold:
i) they  mark the occluded area that needs to be recovered;
ii) they are used to condition a segmentation-aware network that computes optical flows corresponding to the background scene only, 
directly from the occluded input $\burst{}$.
With this network we extract flows $\flowsref{}$ between the keyframe $I_k$ and each other frame $I_j$ in the sequence, and align the burst.
Finally, we employ learned flow-guided image inpainting, to recover the parts of the keyframe that are occluded by the fence, yielding the final output $\keyframe$. 
In the following subsections we explain in detail each  step.

\subsection{Single-frame Fence Segmentation} \label{sec:method:fence_seg}
Our fence segmentation model takes as input a single RGB frame, possibly containing a fence, and outputs a soft fence segmentation mask.
Although this sounds like a relatively simple task, there are, in fact, multiple challenges.
First, datasets of large size and with high quality annotations for fence segmentation are surprisingly scarce.
The most appropriate for this task is probably the De-fencing dataset~\cite{du2018accurate}.
Fences in this dataset do not exhibit significant variance in terms of appearance, scale, or structure, so we rely on substantial data augmentation to train a network that is robust to different types of fences and environments.

More specifically, we apply different degrees of downscaling to the original image and its associated annotation, to effectively create fences at different scales (varying fence width/distance from the camera).
To augment the limited scene variety in the De-fencing dataset, we also mask out fences, using the groundtruth segmentations, and overlay them on images from the DAVIS dataset~\cite{davis}. 
Finally, we apply random horizontal flipping to the fence image and take randomly crop a \res{} window for training.

The segmentation network itself is a \unet~\cite{ronneberger2015u} backbone, with four encoder and four decoder blocks, 
that is trained from scratch on our augmented fence data using a binary cross entropy loss and the ADAM optimizer~\cite{kingma2014adam}.
To obtain segmentation scores in the $[0,1]$ range, we apply a sigmoid in the output logits from the last U-net layer.
\begin{table}[]
    \centering
    \begin{tabular}{|l|c|c|c|}
        \hline
        Method &  Precision & Recall & F-measure \\
        \hline
        Du et al.~\cite{du2018accurate}  &  0.910 &  0.959 &  0.934\\
        \hline
        \unet{} (thresh=0.05) & 0.908 & 0.958 & 0.931 \\
        \hline
        \unet{} (thresh=0.1)  & 0.934 & 0.942 & \textbf{0.937} \\
        \hline
        \unet{} (thresh=0.3)  & 0.969 & 0.899 & 0.932 \\
        \hline
    \end{tabular}
    \caption{Segmentation results on the De-fencing test set~\cite{du2018accurate}.}
    \label{tab:fence_seg}
\end{table}
Table~\ref{tab:fence_seg} lists precision-recall and f-measure scores of our method at different thresholds. 
% Compressible text across these two sentences
Even though we do not use temporal information from multiple frames like~\cite{du2018accurate}, we achieve
comparable performance.

% \begin{figure}
%     \centering
%     \includegraphics[width=\linewidth]{example-image-c}
%     \caption{Our fence segmentation model can accurately segment fences of different sizes and appearance. 
%     \st{maybe we can skip this figure and include segmentation results in the same figure we show inpainting results}}
%     \label{fig:fence_segmentations}
% \end{figure}

\subsection{Segmentation-aware Optical Flow Estimation} \label{sec:method:flow_mask_conditional}
\begin{figure}
    \centering
    \def\wf{0.49}
    \begin{subfigure}{\wf\linewidth}
        \includegraphics[width=\textwidth]{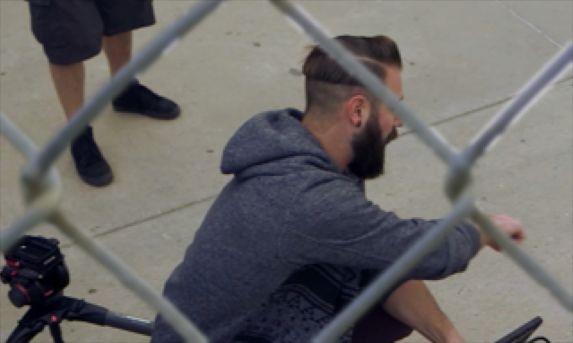}
        \caption{Keyframe}
    \end{subfigure}
    \begin{subfigure}{\wf\linewidth}
        \includegraphics[width=\textwidth]{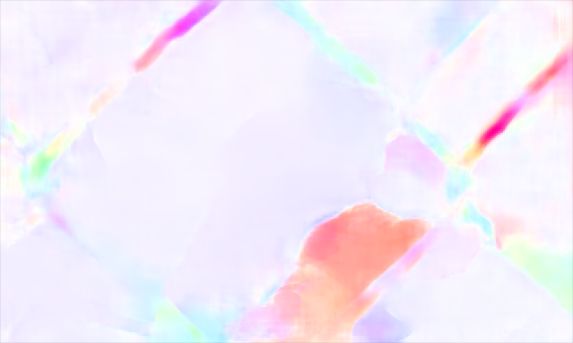}
        \caption{\spynet{}}
    \end{subfigure}
    \begin{subfigure}{\wf\linewidth}
        \includegraphics[width=\textwidth]{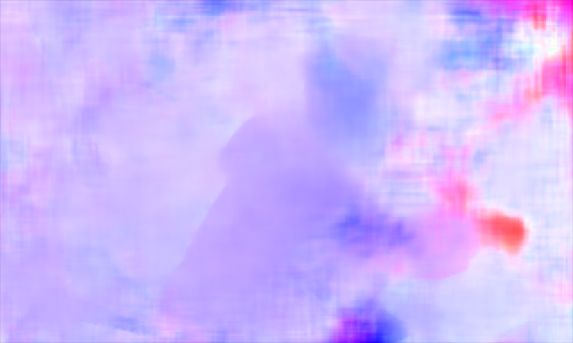}
        \caption{\spynet{}$^m$}
    \end{subfigure}
    \begin{subfigure}{\wf\linewidth}
        \includegraphics[width=\textwidth]{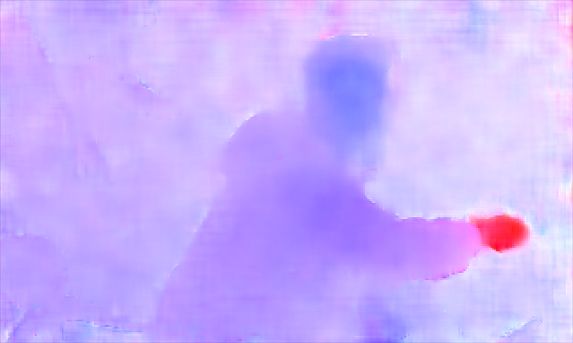}
        \caption{Ground truth}
    \end{subfigure}
    \caption{Standard optical flow networks fail under repeated occlusion patterns. 
    Our occlusion-aware \spynet{}$^m$ can reliably estimate the optical flow \emph{of the background scene}, ignoring the foreground occlusion. 
    For training, ``ground-truth'' flows are computed using a vanilla \spynet{} on the original background frames.}
    \label{fig:mask_conditional_flows}
\end{figure}
Optical flow computation is an integral step in many obstruction removal and video completion pipelines. 
The challenge is in how to align the background regions without being distracted by foreground occlusions.
SOLD~\cite{liu2021learning} uses a pretrained \pwcnet~\cite{sun2018pwc} to compute flows between all frame pairs in the burst, and uses frames warped to the keyframe to prime background reconstruction.
%and warps frames to their corresponding keyframe, priming background reconstruction.
% first estimates a clean reconstruction of the background scene for all the frames in the input sequence. 
% Then, it uses a 
Note that, in the case of fence removal, flows are only computed for the \emph{background} layers, after removing the obstruction.
The reason, according to the authors, is that the flow estimation network cannot handle the repetitive structures, and often predicts noisy results, which renders the alignment step unreliable. 
This is further aggravated by the fact that the weights of \pwcnet{} are frozen, so it cannot adapt to deal with potential 
errors in background layer reconstructions from the first levels in the coarse-to-fine SOLD architecture;
this can consequently yield inaccurate flow estimates in subsequent levels, compounding errors.
% \al{One drawback of SOLD is applying a pretrained PWC-Net to align background reconstructions in intermediate levels. Such reconstructions are not perfect and still contain a mix of foreground and background layers, and may consequently yield inaccurate flow estimates}.
Finally, \pwcnet{} relies on cost volume computation, whose runtime does not scale favorably with input size, 
at least when using a publicly available implementation~\cite{pytorch-pwc}.
On the other hand, \cite{xu2019deep, gao2020flow} compute flow maps between \emph{obstructed} pairs of frames using FlowNet~\cite{ilg2017flownet}.
One key difference in this scenario is that the obstruction does not follow a repetitive structure pattern, but is typically a large, compact area.
This causes the  flow maps to  contain holes which are  inpainted in a separate step. 
% \al {De-fencing paper (Du et al.~\cite{du2018accurate}) also has an occlusion aware flow estimation step, but it involves a lengthy online optimization process.}

In our work we drastically simplify flow estimation for obstructed scenes by utilizing the fence segmentation network described in Section~\ref{sec:method:fence_seg}.
First, we replace \pwcnet{} with the faster, more lightweight, \spynet{}~\cite{ranjan2017optical} architecture\footnote{We use ``SPY'' in equations, for short.}.
Second, we modify its first convolution layer to input both the fence segmentation masks, $S_i$, $S_j$, along with their corresponding input frames $I_i$, $I_j$. Our modified SPY$^m$ architecture then estimates the mask-conditional flow map
% Second, we modify its first convolutional layer, increasing the number of input channels by two (2).
%Given two input frames with fence obstructions $I_i$, $I_j$, we compute their respective fence segmentation maps $S_i$, $S_j$, 
%and pass them as additional inputs to our modified SPY$^m$ architecture to obtain a mask-conditional flow map estimation
\begin{equation}
    f^m_{ij} = \text{SPY}^m([I_i; S_i], [I_j; S_j]),
    \label{eq:mask_conditional_flow}
\end{equation}
$[ \cdot ]$ denoting concatenation along the channel dimension.
We use the original pretrained weights to initialize SPY$^m$, except for the modified part of the input layer, which we initialize randomly.
During training, $f^m_{ij}$ are computed between synthetically generated frames of fence images overlaid on clean background frames.
Consequently, we use flow maps computed with the vanilla \spynet{} on the \emph{clean} background frames $B_i$, $B_j$ as pseudo ground truth targets. 
We use an $L_1$ loss to finetune SPY$^m$:
\begin{equation}
    \mathcal{L}_{f} = \frac{1}{2N}\sum_{\x} | \text{SPY}(B_i, B_j)_{|\x} - f^m_{ij|\x}|,
    \label{eq:mask_conditional_loss}
\end{equation}
where $N$ is the number of image pixels, $\x$ denotes the location at which we evaluate, and we average over $2N$ to account for the $u,v$ flow channels.

Conditioning \spynet{} on segmentation predictions allows us to denote parts of the scene 
corresponding to obstructions and ignore them while computing \emph{background} flows, solving a fundamental problem faced by SOLD.
This idea was previously explored in~\cite{du2018accurate} but it involved a costly optimization process.
Our approach is simple but 
% very effective in capturing motion of the background while being 
robust to the presence of significant fence obstructions. 
Segmentation-aware flow estimation 
% is an interesting problem in its own right, and 
can be useful in a variety of practical
settings where one wants to ignore parts of the scene as distractions or sources of noise.
Figure~\ref{fig:mask_conditional_flows} %validates qualitatively the effectiveness 
demonstrates the effectiveness 
of our approach by comparing outputs of the vanilla \spynet{} and our segmentation-aware \spynet$^m$, on the same obstructed scene. 

% Second, we remove the requirement for frame reconstruction and reliably compute flow maps corresponding to the background scene, directly from obstructed frames.
% Third, we combine flow estimation and inpainting in one single learned prediction, which is both more efficient, and more accurate, as we show in our experiments.
% We achieve this by i) using a light

\subsection{Flow-guided Multi-frame Fence Removal} \label{sec:method:frame_inpainting}
\begin{figure}
    \centering
    \includegraphics[width=\linewidth]{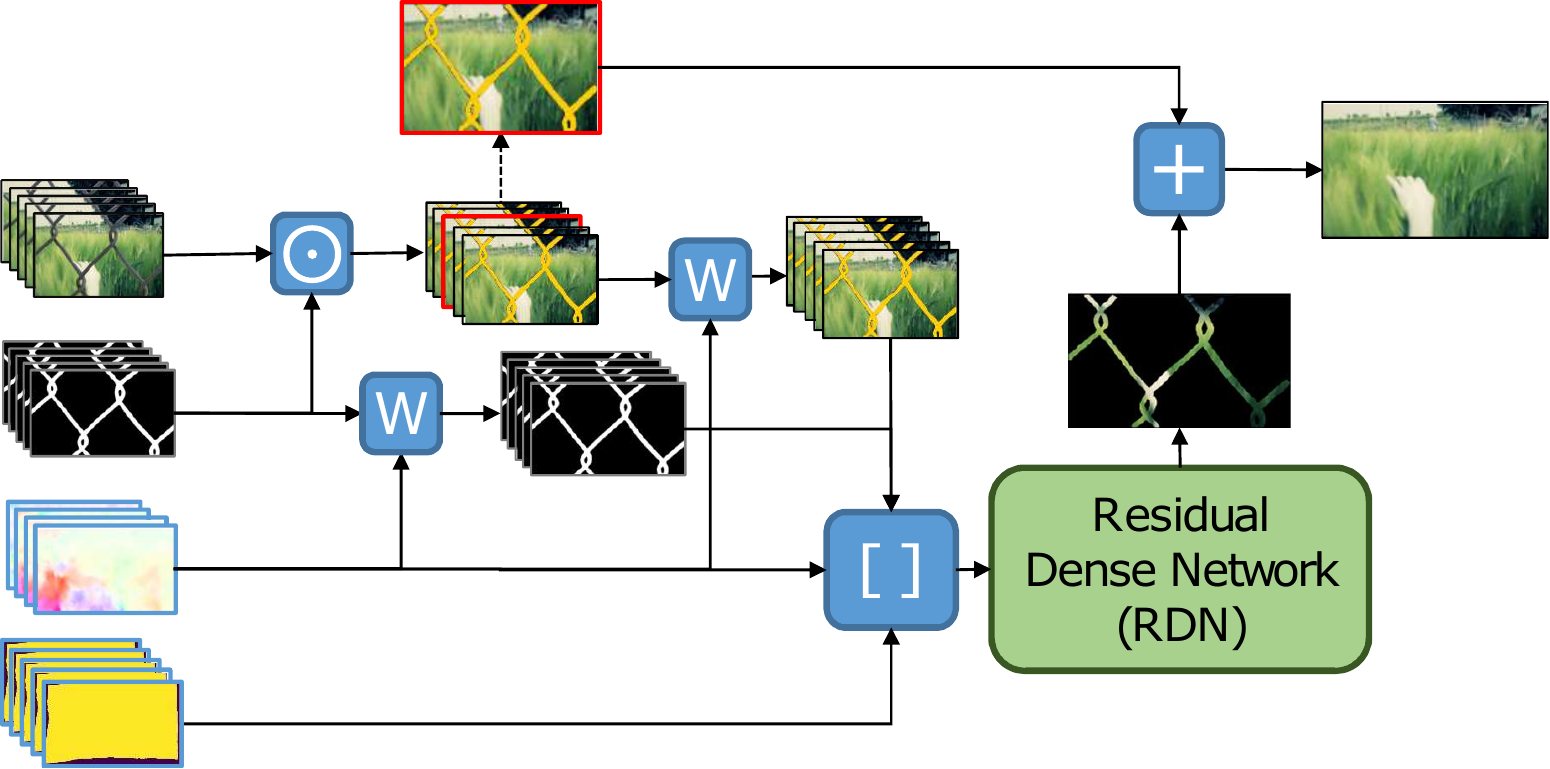}
    \caption{\textbf{Frame inpainting module.}
        We use the predicted fence segmentations to mask out ($\odot$) the occluded areas in the input frames.
        We then use optical flow to warp ($W$) the masked frames and respective masks.
        Finally, the flows, the validity maps (see text), and the aligned frames and masks are concatenated ($[\cdot]$)
        and passed as features to a CNN predicting the keyframe residual in the occluded regions.
    }
    \label{fig:frame_inpainting_module}
\end{figure}
The final component in our fence removal pipeline is a frame inpainting module, depicted in Figure~\ref{fig:frame_inpainting_module}.  % Compression opportunity.
The frame inpainting module takes as input the sequence of obstructed frames $\burst{}$, 
forward flow maps $f^m_{ki}$, computed between the keyframe $I_k$ and each other frame in the burst using the 
mask-conditional \spynet$^m$ (Section~\ref{sec:method:flow_mask_conditional}), and 
the fence segmentation masks $\segmentations{}$ computed using our single-frame segmentation model (Section~\ref{sec:method:fence_seg}).
%Flows $f^m_{ki}$ are used to warp all frames with respect to the reference frame, producing aligned frames $\tilde{I}_i = \text{warp}(I_i, f_{ki})$.
%We also use $\segmentations{}$ to mask out the areas that correspond to fences in each frame, obtaining $\burstmasked{}$.
%We note that, although there are fences in all frames in the sequence, our goal is to inpaint the missing (masked out) areas \emph{only} in the reference frame.
We first use $\segmentations{}$ to mask out the areas that correspond to fences in each frame, obtaining masked frames $I^m_i = I_i \odot S_i$.
Then flows $f^m_{ki}$ are used to warp all frames and their respective segmentations with respect to the reference frame, 
giving rise to aligned masked frames $\tilde{I}^m_i = W(I^m_i, f^m_{ki})$ and aligned fence masks $\tilde{S}_i = W(S_i, f^m_{ki})$. 
We also compute binary masks $\valid{}$ that mark valid warped regions, and are ``on'' for all pixels that fall inside the image grid after warping. 

% \paragraph{Pixel propagation.} 
% After alignment, some of the locations that are occluded in the keyframe are visible in other frames in the sequence, 
% so we can directly ``propagate'' their RGB values, similarly to~\cite{xu2019deep}.
% This not only reduces the number of pixels values that have to be inpainted, but also offers realistic context for the inpainting process.
% We estimate these locations as a soft mask union array
% \begin{equation}
%     \{ M^u_i \}: M^u_i = S_k \cdot (1 - \tilde{S}^m_i) \cdot V_i.
%     \label{eq:mask_union}
% \end{equation}
% Because the occluded pixel may be visible in more than one frames in the sequence, we weigh the contributions of these different values
% using their respective fence confidence scores from $M^u_i$, so the propagated pixel values in the occluded area are
% \begin{equation}
%     P = \frac{\sum M^u_i \cdot \tilde{I}^m_i}{\sum M^u_i + \epsilon} 
%     \label{eq:pixelprop_weight}
% \end{equation}
% and the initialized keyframe after pixel propagation is
% \begin{equation}
%     I^P_k = I_k \cdot (1 - S_k) + S_k \cdot V_k \cdot P.
%     \label{eq:pixelprop_keyframe}
% \end{equation}

% \paragraph{Frame inpainting.}
% $f_{in} = [ I^P_k; \{\tilde{I}^m_i\};  \tilde{\segmentations{}}; \valid{} ]$ 
$f_{in} = [ \{\tilde{I}^m_i\};  \tilde{\segmentations{}}; \valid{} ]$ 
is passed as input to a Residual Dense Network (RDN)~\cite{zhang2018residual} that is responsible for filling in the missing areas in the keyframe. 
We also add a skip connection between the masked keyframe $I^m_k$ and the output of the RDN, so the latter only has to 
learn to fill in the missing areas instead of reconstructing the entire image.
The inpainting module is trained in a supervised fashion using an $L_1$ loss and the clean background as the ground truth:
\begin{equation}
    \mathcal{L}_{in} = \frac{1}{N}\sum_{\x} |B_{k|\x} - (I^m_k + \text{\fim{}}(f_{in}))_{|\x}|.
    \label{eq:loss_inpaint}
\end{equation}

\subsection{Implementation details} \label{sec:method:implementation}
We implement our pipeline in Python 3 and PyTorch~\cite{pytorch}.
For \unet{}, \pwcnet{}, \spynet{}, and RDN, we use their publicly available 
third-party implementations~\cite{pytorch-unet, pytorch-spynet, pytorch-pwc, pytorch-rdn}. 
To facilitate our experiments, we have also re-implemented SOLD in PyTorch (\soldpt{}),
following closely the original Tensorflow implementation~\cite{liu2021learning}; 
we plan to make our re-implementation publicly available to allow for broader use by the community and replication of results.
Unless otherwise mentioned, we train all our models for 1000 epochs, using a starting learning rate $l_r=10^{-4}$, a weight decay 
rate $w_r = 4\cdot10^{-5}$, and the ADAM optimizer~\cite{kingma2014adam} with 
parameters $\alpha=10^{-4}$, $\beta_1=0.9$, $\beta_2=0.999$, $\epsilon=10^{-8}$.
All three models (fence segmentation, occlusion-aware flow estimation, frame inpainting)
are trained independently.

% \paragraph{Frame inpainting module.}
% We have experimented with two types of convolutional networks:
% i) a simple 5-layer convolutional network \st{add details}, which is in fact the same network used for background and foreground layer reconstruction in~\cite{liu2021learning};
% ii) a Residual Dense Network~\cite{zhang2018residual}.

\section{Data for Training and Evaluation} \label{sec:data}
We use two types of data in our experiments. 
The first type is synthetic multi-frame sequences, generated similarly to  previous works~\cite{du2018accurate,liu2021learning}. 
These are used predominantly for training and validation experiments, but a held-out test set is also used for evaluation.
The second type is real bursts with fence obstructions, which include \emph{uncontrolled} sequences, 
for which no ground truth clean frame is available, and \emph{controlled} sequences, which come with a clean background scene (without the fence) as ground truth.

% \paragraph{Synthetic bursts} 
\medskip
\noindent\textbf{Synthetic bursts} 
are generated by overlaying obstruction (foreground) layers on a clean scene (background).
% Because this step can critically impact the performance of the de-fencing model, we provide details of our synthetic sequence generation process.
We source background scenes (which are also used as ground truth during training and evaluation) from 
Vimeo-90k~\cite{vimeo90k}, which consists of videos depicting every day activities in 
realistic settings, often including people and other objects.
We specifically use the original test split of the dataset\footnote{\small \url{http://data.csail.mit.edu/tofu/testset/vimeo_test_clean.zip}}, 
which contains sequences of seven (7) frames.
Training and validation splits are generated on the fly, but for our evaluation experiments we use a fixed test set of 100 bursts.
% , as the background scenes on which we overlay fence obstructions. 
% The original clean frames are used as ground truth for training and evaluation.
% To increase variability of our synthetically generated data, we apply the following data augmentation steps:
% a) random homography transformation;
% b) center cropping to avoid any black borders caused by a);
% c) random cropping of a $192\times320$ window, which are the frame dimensions used during training;
% d) random horizontal flip.

The foreground fence obstructions are sourced from the De-fencing dataset~\cite{du2018accurate}, 
which contains 545 training and 100 test images with fences, along with corresponding binary masks as ground truth for the fence segmentation.
The sequences in this dataset have been collected in various outdoor conditions and have a variable frame count per scene.
Since we have the ground truth fence masks, we can use them to mask out the fence from any given frame and overlay it on
a clean background from Vimeo.
To obtain a fence image burst of size $K$, we mask out the fence from a single frame and apply $K$ random 
perspective distortions to it, to simulate the changes caused by slightly different viewpoints and motion.
To increase the variability of fences and background scenes, we apply various forms of data
augmentations before fusing them into a single frame; these are listed in detail in the supplemental material.

% However, since variability of fences is limited, we also apply various forms of data augmentation on the fence image
% before fusing it with the background.
% The types of foreground augmentation we consider are:
% a) random downsample of the fence image and segmentation to create fences of different sizes and thickness;
% b) random ``outer'' window crop to focus on a specific subregion of the fence;
% c) color jitter to make the network more robust to different fence appearances and lighting conditions;
% d) random perspective distortion to obtain a fence sequence of length $K$;
% e) center cropping to avoid black border effects from the homographic distortion;
% f) random blur with a gaussian kernel, to simulate defocus aberrations.

% \paragraph{Real bursts.}
\medskip
\noindent\textbf{Real bursts.}
Because we want to develop a practical algorithm for fence removal, good performance under realistic
motion, lighting, and obstruction patterns is of paramount importance.
In previous works, performance on real sequences is --for the most part-- evaluated qualitatively, 
since obtaining the ground truth background is far from trivial.
Liu et al.~\cite{liu2021learning} include only two sequences with fence-like obstructions, collected in a controlled environment, 
which is too small a dataset for a proper quantiative evaluation.

In this paper we construct a wider set of controlled sequences, specifically for quantitative evaluation. 
Rather than collecting toy scenes as in Liu et al.~\cite{liu2021learning}, we capture real world hand-held sequences 
with a fence and a corresponding background ground truth image without a fence.
%using the IPhone11Pro-Max. 
As we cannot physically remove a fence, we instead bring our camera to the fence and center it in one of the fence cells such that only the background is visible. 
To maintain a similar level of brightness and sharpness of the background in the input and ground truth images, 
we fix the exposure and focus of the camera on the background during capture. 
Due to camera motion and possible changes in illumination, the input keyframe and its respective ground truth may be misaligned or have color discrepancies. 
We align crops of the scene using standard feature-based RANSAC fitting of homographies, similar to \cite{bhat2021deep} and 
% to the method used for the real BurstSR dataset of Bhat et al.~\cite{bhat2021deep}. 
correct color discrepancies using color histogram matching. 
We then filter out any misaligned crops using SSIM, PSNR, and human visual check, avoiding 
mostly homogeneous regions, to promote diversity in our dataset. 
Our final real burst dataset consists of $185$ \res{} input bursts and corresponding ground truth keyframes. 
More details on dataset generation and image sampling are provided in supplemental material.

\section{Experiments} \label{sec:experiments}
We compare our method and other works on synthetic and real bursts.
For quantitative evaluations we use the test set of our synthetically generated fence-obstructed sequences,
and our real bursts described in Section~\ref{sec:data}.
For all baselines, we use the officially released model weights, with the exception of our SOLD reimplementation.
We also purposedly omit the sequence-specific online optimization step of SOLD in our comparisons.
Although online optimization improves performance, its runtime is quite slow ($\sim 3$  minutes per burst), 
pushing it outside the scope of our work, which is centered around efficiency and practicality.
% Besides, we could also employ a similar meta-learning procedure to improve the performance of our approach.  
For qualitative evaluations and visual comparison, we use real sequences from previous work and the data we collected.

\subsection{Baselines} \label{sec:experiments:baselines}
% \paragraph{Single-frame baseline.}
\noindent\textbf{Single-frame baseline.}
We pass either ground truth or our (thresholded) \unet{} fence mask predictions as input to LaMa~\cite{suvorov2022resolution}, 
a \sota{} CNN-based inpainting method, to create a single-frame de-fencing baseline.
LaMa takes as input a (possibly obstructed) image and a binary mask and inpaints the area marked by the mask.
\medskip

% \paragraph{SOLD~\cite{liu2021learning}} 
\noindent\textbf{SOLD~\cite{liu2021learning}} 
primarily targets reflection removal, 
but it can be adapted to deal with opaque obstructions such as fences or raindrops on glass.
We evaluate both the original Tensorflow model (\soldtf{}) and our PyTorch reimplementation (\soldpt{}), with the latter trained on our synthetic data.
\medskip

% \paragraph{Flow-guided video completion} 
\noindent\textbf{Flow-guided video completion} 
operates in a setting that is different than ours in a few  ways.
First, the mask denoting the occluded area is known, and its shape is either rectangular or in the shape of 
an object in the video.
Second, the number of frames in a typical input video sequence is $K \gg 5$. 
Lastly, the output is the entire inpainted video.
Nevertheless, we can apply these methods for de-fencing in a relatively straightforward fashion by 
passing the fence segmentation as the occlusion mask, treating the burst as a (short)
video sequence, and keeping the inpainted result for the reference frame only.
In our experiments we compare against two recent flow-guided approaches, FGVC~\cite{gao2020flow}
and E2FGVI~\cite{li2022towards}, using their publicly provided code.

\subsection{Fence Removal on Synthetic and Real Data} \label{sec:experiments:results}
\begin{figure*}[t]
    \def\wf{0.16}
    \centering
    \includegraphics[width=0.98\linewidth]{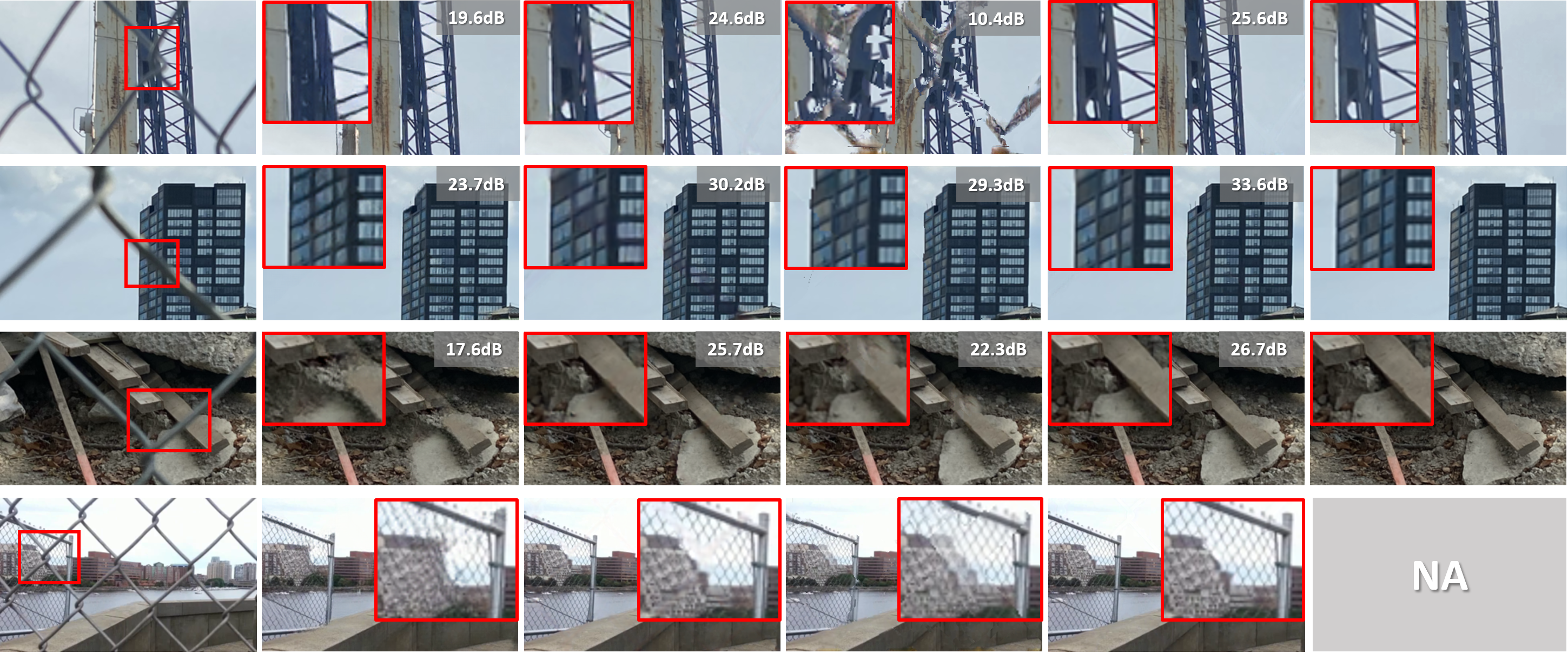}
    \begin{subfigure}{\wf\textwidth}
        \caption{Keyframe}
    \end{subfigure}
    \begin{subfigure}{\wf\textwidth}
        \caption{LaMa~\cite{suvorov2022resolution}}
    \end{subfigure}
    \begin{subfigure}{\wf\textwidth}
        \caption{SOLD~\cite{liu2021learning}}
    \end{subfigure}
    \begin{subfigure}{\wf\textwidth}
        \caption{FGVC~\cite{gao2020flow}}
    \end{subfigure}
    \begin{subfigure}{\wf\textwidth}
        \caption{Ours}
    \end{subfigure}
    \begin{subfigure}{\wf\textwidth}
        \caption{Ground truth}
    \end{subfigure}
    \caption{Qualitative de-fencing results on real sequences. We highlight areas of interest in red (shown as zoomed insets) 
        and report PSNR inside the fence mask for all methods. Last example is from Xue et al.~\cite{xue2015computational} and has no ground truth.
    }
    \label{fig:frame_inpainting_qualitative}
\end{figure*}

\begin{table*}
    \newcommand{\blue}[1]{\textcolor{blue}{#1}}
    \resizebox{\textwidth}{!}{
        \begin{tabular}{|l|c|c|c|c|c|c|c|}
            \hline
            \multicolumn{8}{|c|}{\textbf{Synthetic data} (Methods in \textcolor{blue}{blue} use GT fence masks)} \\
            \hline
            \multirow{2}{*}{Method}  & \multicolumn{3}{|c|}{SSIM $\uparrow$ } & \multicolumn{3}{|c|}{PSNR (dB) $\uparrow$} & LPIPS $\downarrow$ \\
            \cline{2-7}
             & in & out & total & in & out & total & (VGG) \\
            \hline
             \soldtf{}\cite{liu2021learning}                & .783 & .970 & .941 & 23.36 & 37.82 & 30.34 & .111 \\
            \hline
             \soldpt{}~\cite{liu2021learning}               & .893 & .993 & .977 & 28.14 & 45.70 & 35.82 & .040 \\
            \hline
             \blue{LaMa}~\cite{suvorov2022resolution}       & .788 & .995 & .964 & 24.97 & 51.38 & 33.10 & .039 \\
            \hline
             \blue{LaMa}~\cite{suvorov2022resolution}$^*$   & .655 & .955 & .910 & 20.96 & 31.74 & 27.38 & .089 \\
            \hline
             \blue{FGVC}~\cite{gao2020flow}                 & .846 & .943 & .928 & 25.56 & 33.85 & 30.36 & .068 \\
            \hline
             \blue{FGVC}~\cite{gao2020flow}$^*$             & .784 & .896 & .879 & 22.73 & 27.80 & 26.05 & .113 \\
            \hline
             \blue{E2FGVI}~\cite{li2022towards}             & .918 & .997 & .985 & 30.79 & 55.87 & 38.89 & .030 \\
            \hline
             \blue{E2FGVI}~\cite{li2022towards}$^*$         & .890 & .984 & .969 & 29.34 & 38.64 & 35.25 & .044 \\
            \hline
             Ours                                           & \textbf{.954} & \textbf{.999} & \textbf{.992} & \textbf{33.76} & \textbf{56.55} & \textbf{41.78} & \textbf{.015} \\
            \hline
            \hline
             \blue{Ours-fence$_{gt}$}                       & \emph{.957} & \emph{.999} & \emph{.993} & \emph{34.33} & \emph{58.58} & \emph{42.42} & \emph{.012} \\
            \hline
        \end{tabular}
    
        \begin{tabular}{|c|c|c|c|c|c|c|}
            \hline
            \multicolumn{7}{|c|}{\textbf{Real bursts} (Methods in \textcolor{blue}{blue} use \emph{pseudo}-GT fence masks)} \\
            \hline
             \multicolumn{3}{|c|}{SSIM $\uparrow$ } & \multicolumn{3}{|c|}{PSNR (dB) $\uparrow$} & LPIPS $\downarrow$ \\
            \cline{1-6}
             in & out & total & in & out & total & (VGG) \\
            \hline
             .728 & .911 & .885 & 23.50 & 30.27 & 27.71 & .132 \\
            \hline
             .813 & .916 & .902 & 26.41 & 30.90 & 29.71 & .094 \\
            \hline
             .480 & .902 & .845 & 19.95 & 29.96 & 26.25 & .133 \\
            \hline
             .477 & .867 & .816 & 20.85 & 28.02 & 25.98 & .132 \\
            \hline
              .848 & .910 & .901 & 27.56 & 30.48 & 29.73 & .095 \\
            \hline
             .856 & .907 & .900 & 27.49 & 29.90 & 29.36 & .090 \\
            \hline
             .571 & .902 & .856 & 19.69 & 29.88 & 25.87 & .167 \\
            \hline
             .709 & .901 & .875 & 25.58 & 30.25 & 29.14 & .117 \\
            \hline
             \textbf{.869} & \textbf{.917} & \textbf{.909} & \textbf{28.60} & \textbf{31.14} & \textbf{30.46} & \textbf{.080} \\
            \hline
            \hline
             \emph{.872} & \emph{.918} & \emph{.910} & \emph{28.77} & \emph{31.15} & \emph{30.53} & \emph{.078} \\
            \hline
        \end{tabular}
    }
    \caption{Results on synthetic test data (left) and our collected real bursts (right). 
        Results in rows denoted with ``*'' are computed after thresholding the fence segmentations at $t=0.1$, 
        and dilating the binary mask 4 times with a $3\times3$ square.
    }
    \label{tab:results}
\end{table*}

Quantitative comparisons on synthetic and real bursts are shown in Table~\ref{tab:results}.
We report performance in terms of the commonly used SSIM, PSNR, and LPIPS metrics. 
For LPIPS we use a VGG-16 backbone as the feature extractor.
PSNR and SSIM can be computed as an aggregation of pixel-wise scores, so we use the fence masks 
(ground truth in the case of synthetic data, thresholded and binarized \unet{} predictions in the case of real data\footnote{To get better pseudo ground truth fence masks, we run \unet{} at multiple scales and compute
the pixel-wise maximum across scales.})
to dissect performance in three different regions: a) inside the mask (in); b) outside the mask (out); and c) in the entire image (total).
Performance outside the mask is high for all methods, since this part of the image is not occluded.
Performance \emph{inside} the mask is the most important criterion, since it quantifies the quality of reconstruction
only in the occluded area.
We outperform all single- and multi-frame baselines, based on all metrics, with FGVC~\cite{gao2020flow}
performing second best at the cost of much higher runtime (see Section~\ref{sec:experiments:runtime}).
We would like to draw the reader's attention to the results of LaMa~\cite{suvorov2022resolution}, in particular.
LaMa is a \sota{} inpainting method, yet it achieves surprisingly low PSNR-in and SSIM-in scores. 
The reason becomes clear if one looks at Figure~\ref{fig:frame_inpainting_qualitative} (e.g., antennae structure in the first example):
even though LaMa produces perfectly plausible results under the occluded area,
these are often very different than the actual background scene.
These results tellingly demonstrate the advantage of using multiple frames for de-fencing. 
Figure~\ref{fig:frame_inpainting_qualitative} also illustrates that alternatives
like SOLD and FGVC can yield blurry or completely scrambled results, likely due to issues with frame alignment. 
For more qualitative results see our supplemental material.

\subsection{Runtime Analysis} \label{sec:experiments:runtime}
\begin{table}
    \centering
    \resizebox{\linewidth}{!}{
    \begin{tabular}{|c|c|c|c|c|c|}
        \hline
         Method & LaMa & \soldpt{} & FGVC & E2FGVI &  Ours  \\
        %  Method & LaMa~\cite{suvorov2022resolution} & \soldpt{}~\cite{liu2021learning} & FGVC~\cite{gao2020flow} &E2FGVI~\cite{li2022towards} &  Ours  \\
        \hline
         Runtime (s) & 0.2 & 0.8 & 0.7 & 0.16 & \textbf{0.14} (\textbf{0.08})   \\
        \hline
    \end{tabular}}
    \caption{Runtime comparison for a 5-frame burst
        % (except LaMa, which processes a single frame). 
        (LaMa processes a single frame). 
        Times for other methods \emph{do not} include segmentation 
        (fence segmentations are part of the SOLD output).
        We provide the runtime of our method without the segmentation step in parentheses, for comparison.
    }
    \label{tab:runtime}
\end{table}

% Table~\ref{tab:runtime} compares the total runtime of our pipeline with other approaches.
% In our case, timing includes all necessary processing steps: fence segmentation, optical flow computation, alignment, and frame inpainting. 
% Note that LaMa, FGVC, and E2FGVI times \emph{do not} include the time spent on fence segmentation.
% Our method is clocked at $\sim \fps{}$, for \res{} inputs, which is comparable to E2FGVI and LaMa; 
% however, we outperform both approaches on synthetic and real data,
% and we are also  $\sim 5\times$ faster than the next best performing methods, SOLD and FGVC.
% All timings are performed on a high end workstation equipped with a Nvidia GTX 1080 Ti, with 12GB of GPU RAM.

Table~\ref{tab:runtime} compares the total runtime of our pipeline with other approaches.
In our case, timing includes all necessary processing steps: fence segmentation, optical flow computation, alignment, and frame inpainting. 
LaMa, FGVC, and E2FGVI times \emph{do not} include the time spent on fence segmentation, 
since these methods assume that the occlusion masks are precomputed.
Our method is clocked at $\sim \fps{}$, for \res{} inputs, which is $\sim 5\times$ faster than the next best performing methods, SOLD and FGVC, 
and comparable to E2FGVI and LaMa.
The runtime difference with the last two may not seem large in absolute terms, 
but it still amounts to a $12.5\%$ and $30\%$ lower runtime respectively, 
while our method \emph{significantly} outperforms them in terms of reconstruction quality on real data.
If we exclude the time spent on segmentation from our pipeline, the speedup becomes even more noticeable ($50\%$ and $60\%$ respectively).
All timings are performed on a workstation equipped with a Nvidia GTX 1080 Ti, with 12GB of GPU RAM.
The detailed breakdown of timings for our pipeline is:
i) segmentation (for a 5-frame burst): $0.06s$;
ii) flow estimation and alignment: $0.04s$;
iii) frame inpainting (RDN): $0.04s$.

\subsection{Ablation} \label{sec:experiments:ablation}

% \paragraph{Explicit pixel propagation}using \eqref{eq:pixelprop_keyframe} offers a good initialization for the occluded area in the keyframe,
% reducing the number of RGB values
% that must be hallucinated by the RDN.
% Removing this initialization decreases performance by \st{add dB}.

% \paragraph{Changing the inpainting module architecture} 
\noindent\textbf{Changing the frame inpainting module architecture} 
allows us to trade-off reconstruction performance for efficiency.
Replacing the RDN with a simple CNN consisting of 8 convolution + LeakyReLU layers
% (identical to the frame reconstruction module from~\cite{liu2021learning}),
decreases performance by \textcolor{red}{3.5 dB} on synthetic test data but also reduces runtime by $\sim 30\%$,
from $0.14s$ to $0.1s$.
\medskip

% \paragraph{Does frame alignment matter?}
\noindent\textbf{Does frame alignment matter?}
Inaccurate computation of the motion corresponding to the (occluded) background can lead to errors in frame alignment,
impacting the quality of frame reconstruction.
We experiment with the following two alternatives for computing flows:
i) original \spynet{} on obstructed frames;
ii) \spynet{} followed by masking the occluded areas and using 
Laplacian inpainting~\cite{xu2019deep} to complete the missing flows (\spynet{}$^{inp}$).
We also consider the option of not aligning frames at all, and leaving our RDN inpainting network to learn how to complete the 
occluded areas in the reconstructed keyframe.

The importance of alignment and the quality of flows used become
becomes clear when looking at the results in Table~\ref{tab:flow_ablation}.
Not aligning the input frames at all leads to a noticeable drop in keyframe reconstruction performance.
Standard flow networks cannot handle the repeated 
fence obstruction patterns, and explicitly inpainting the flows under the occluded area does not help
either since flow artifacts extend to non-fence areas as well, as shown in Figure~\ref{fig:mask_conditional_flows}.
Our occlusion-aware \spynet$^m$, on the other hand, can accurately estimate background flows,
resulting in superior frame alignment and reconstruction quality.

\begin{table}[]
    \centering
    \resizebox{\linewidth}{!}{
        \begin{tabular}{|l|c|c|c|c|c|c|c|}
            \hline
            \multirow{2}{*}{Alignment}  & \multicolumn{3}{|c|}{SSIM $\uparrow$ } & \multicolumn{3}{|c|}{PSNR (dB) $\uparrow$} & LPIPS $\downarrow$ \\
            \cline{2-7}
             & in & out & total & in & out & total & (VGG) \\
            \hline
            None                    & .792 & .996 & .965 & 24.82 & 53.64 & 32.84 & .028 \\
            \hline
             \spynet{}              & .841 & .997 & .973 & 26.84 & 54.32 & 34.93 & .047 \\
            \hline
             \spynet{}$^{inp}$      & .841 & .997 & .973 & 26.81 & 54.13 & 34.89 & .048 \\
            \hline
             \spynet{}$^m$          & \textbf{.954} & \textbf{.999} & \textbf{.992} & \textbf{33.76} & \textbf{56.55} & \textbf{41.78} & \textbf{.015} \\
            \hline
        \end{tabular}
    }
    \caption{Effect of frame alignment on keyframe reconstruction quality (results on our synthetic test data).
        % Inpainting \spynet{} flows in the occluded area does not fix errors
        % caused because of the repeated obstruction pattern.
        % Our occlusion-aware \spynet{}^$m$ solves this problem
    }
    \label{tab:flow_ablation}
\end{table}

\subsection{Limitations and Failure Cases.} \label{sec:experiments:limitations}
\begin{figure}
    \centering
    \includegraphics[width=0.32\linewidth]{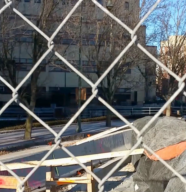}
    \includegraphics[width=0.32\linewidth]{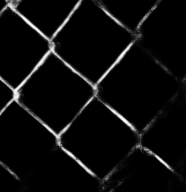}
    \includegraphics[width=0.32\linewidth]{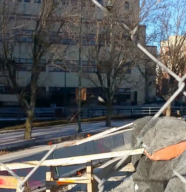}
    \caption{\textbf{Failure example.} Imperfect fence predictions (middle) compromise the quality of the inpainting (right).}
    \label{fig:failures}
\end{figure}

The main limitation of our approach is that the quality of the final reconstruction 
depends on the outputs of \unet{} and \spynet{}$^m$ in the two previous stages.
Errors in fence segmentation affect segmentation-aware optical flow computation, 
potentially compromising frame alignment, which is crucial for good inpainting (see ablation in Section~\ref{sec:experiments:ablation}). 
In addition, fence segmentations are also used in masking out the occluded areas in $\burst{}$, 
The most deleterious mistakes occur when the fence occlusion is out of our training distribution, e.g., 
when the fence has an unusual shape/pattern, or when the contrast with the background is low. 
This is an issue that can be handled to a certain extent through better data augmentation during training
or by having access to richer datasets with varied types of fences, as we show in the supplemental material. 
% Handling different types of fences is something we plan to address in future work.
A failure example and its effect on frame inpainting is shown in Figure~\ref{fig:failures}.

\section{Conclusions} \label{sec:conclusions}
We have developed a simple, modular, and efficient pipeline for removing fence obstructions
from a singe frame in a photo burst.
Our algorithm enjoys the realism of flow-guided video completion methods, while addressing some 
of their practical limitations, such as complicated training and long runtimes. 
Our method runs at $\fps{}$ for 5-frame \res{} bursts, on a Nvidia GTX 1080 Ti, 
and is particularly effective on \emph{real} data, 
outperforming other single- and multi-frame de-fencing baselines 
on a dataset of obstructed bursts we collected specifically for this problem.

%------------------------------------------------------------------------

{\small
\bibliographystyle{ieee_fullname}
\bibliography{egbib}
}

\newpage
% \clearpage
\appendix
\section{Data}

\subsection{Synthetic Data Augmentation}
\begin{figure*}[t!]
    \centering
    \includegraphics[width=\textwidth]{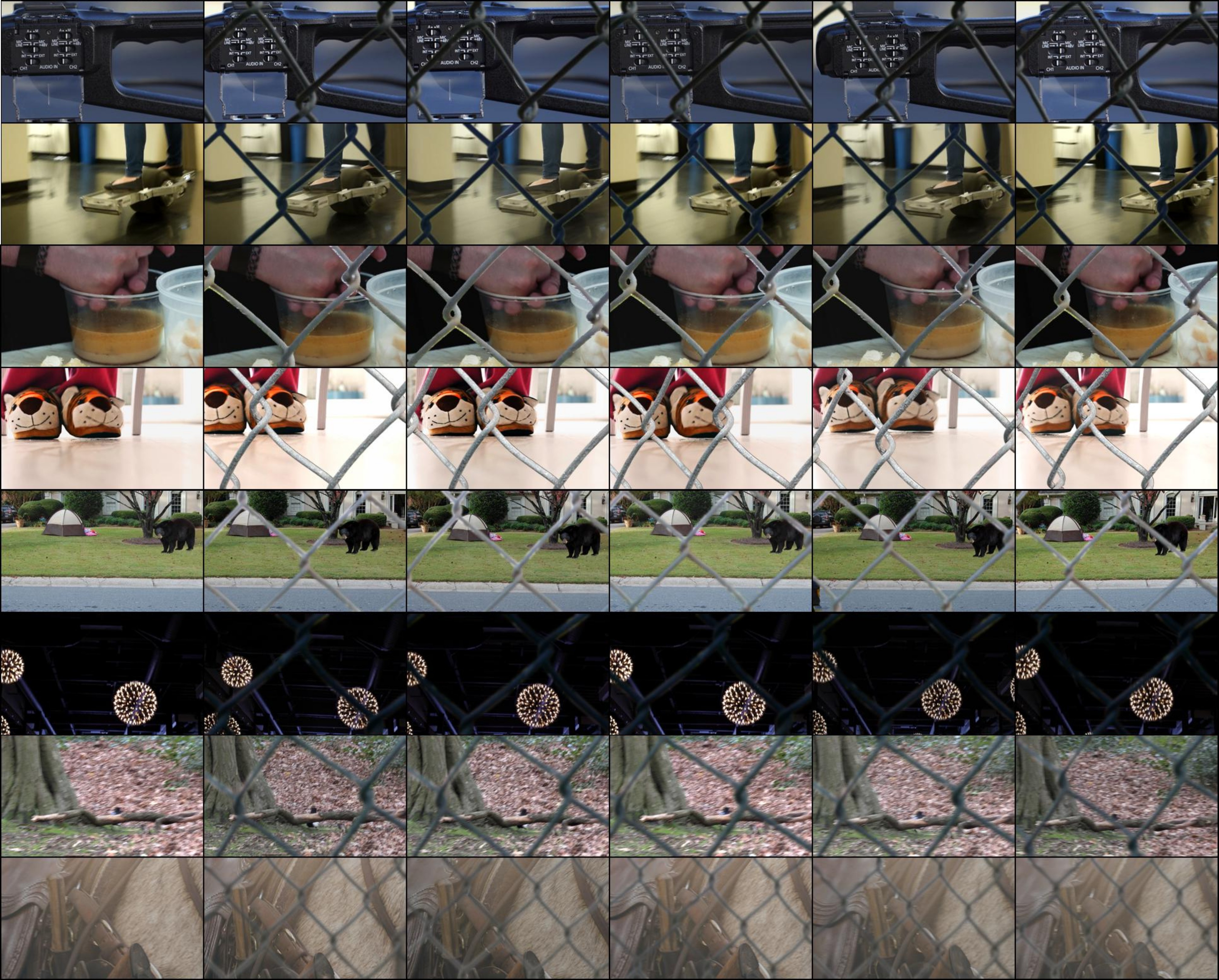}
    \caption{Examples of our synthetically generated data. 
    The leftmost column shows the clean background frame and    the next 5 columns show the background burst from 
    Vimeo-90k~\cite{vimeo90k}, with overlaid fences from the De-fencing dataset~\cite{du2018accurate}.}
    \label{fig:sythetic_data}
\end{figure*}
\paragraph{Background augmentation.} 
We source background scenes (which are also used as ground truth during training and evaluation) from 
Vimeo-90k~\cite{vimeo90k}, which consists of videos depicting every day activities in 
realistic settings, often including people and other objects.
We specifically use the original test split of the dataset\footnote{\url{http://data.csail.mit.edu/tofu/testset/vimeo_test_clean.zip}}, 
which contains sequences of seven (7) frames.
The original clean frames are used as ground truth for training and evaluation.
To increase variability of our synthetically generated data, we apply the following data augmentation steps:
\begin{enumerate}
    \item random homography transformation
    \item center cropping to avoid any black borders caused by (1).
    \item random cropping of a \res{} window, which are the frame dimensions used during training.
    \item random horizontal flip.
\end{enumerate}
% a) random homography transformation;
% b) center cropping to avoid any black borders caused by a);
% c) random cropping of a \res{} window, which are the frame dimensions used during training;
% d) random horizontal flip.

\paragraph{Foreground augmentation.} 
The foreground fence obstructions are sourced from the De-fencing dataset~\cite{du2018accurate}, 
which contains 545 training and 100 test images with fences, along with corresponding binary masks as ground truth for the fence segmentation.
The variability of fences in that dataset is limited, 
so we also apply various forms of data augmentation on the fence image before fusing it with the background.
The types of foreground augmentation we consider are:
\begin{enumerate}
    \item random downsample of the fence image and segmentation to create fences of different sizes and thickness.
    \item random ``outer'' window crop to focus on a specific subregion of the fence.
    \item color jitter to make the network more robust to different fence appearances and lighting conditions.
    \item random perspective distortion to obtain a fence sequence of length $K$.
    \item center cropping to avoid black border effects from the homographic distortion.
    \item random blur with a gaussian kernel, to simulate defocus aberrations.
\end{enumerate}
% a) random downsample of the fence image and segmentation to create fences of different sizes and thickness;
% b) random ``outer'' window crop to focus on a specific subregion of the fence;
% c) color jitter to make the network more robust to different fence appearances and lighting conditions;
% d) random perspective distortion to obtain a fence sequence of length $K$;
% e) center cropping to avoid black border effects from the homographic distortion;
% f) random blur with a gaussian kernel, to simulate defocus aberrations.
Samples from our synthetic burst dataset are shown in Figure~\ref{fig:sythetic_data}.

\subsection{Real Burst Collection}
Although our synthetic data are carefully generated and exhibit considerable realism and diversity, they still cannot
fully capture the variability of motion, lighting, and obstruction patterns in scenes captured under realistic conditions, 
so we collect set of controlled sequences, specifically for quantitative evaluation. 
As mentioned in the main paper, rather than collecting toy scenes as in Liu et al.~\cite{liu2021learning}, 
we capture outdoors real world hand-held sequences with a fence and a corresponding background ground truth image without a fence.

\paragraph{Data capture.}
We first capture one image without the fence as the ground-truth frame, by bringing our camera to the centre of a fence cell. 
We then fix the focus and exposure on the background and move backwards from the fence to capture 5 frames with fences.  
To minimize misalignment caused by a change in perspective, we capture the first frame as the key-frame, moving backwards along the capturing direction. 
Then, we capture the remaining four frames by intentionally jittering the camera around.

\begin{figure*}
   \centering
    \def\wf{0.19}
    \centering
    \includegraphics[width=\textwidth]{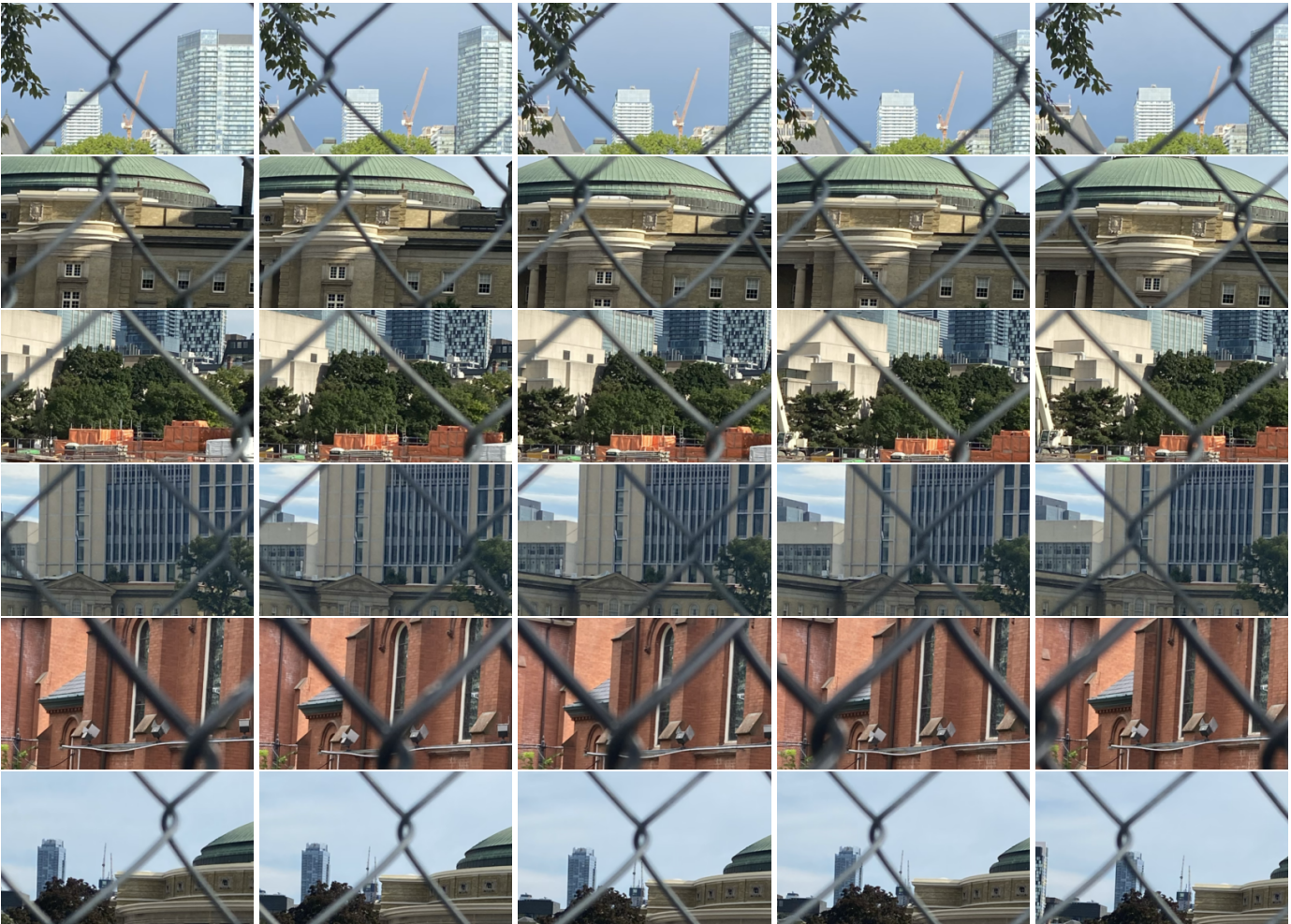}
    \caption{Examples of real bursts we have collected. These are \res{} crops from the original, high resolution images, after alignment.}
    \label{fig:real_bursts}
\end{figure*}

\paragraph{Keyframe - groundtruth alignment.}
After capturing the real bursts, we need to align the ground-truth frame to the obstruced key-frame. 
We do this following an approach combining SIFT feature extraction and RANSAC homography estimation, similar to  \cite{bhat2021deep}.
We start by computing and matching SIFT features in the keyframe and respective clean groundtruth shot.
Since the resolution of the original images is high, we extract \res{} regions in a sliding window fashion,
and within such window, $P$, we compute homography parameters using matched SIFT features in crops of varying sizes: 
$128^2$, $256^2$, $512^2$, and $1024^2$ (larger crop sizes extend beyond the area of the original window).
The computed homography parameters are used for global alignment of the keyframe and groundtruth frames, so
we have multiple homography ``candidates'' corresponding to $P$.
The motivation behind computing homographies at different scales is that 
different parts of a given window $P$ may require different homographies to be aligned more accurately.
We assign the best homography to each $128\times128$ crop $C$ inside $P$, by computing
its respective SSIM score with respect to its warped counterpart in the groundtruth (we use the estimated fence masks
to only include non-obstructed areas in the SSIM computation).
To ensure a minimum level of quality, if there is at least one $C$ inside $P$ with average SSIM $\leq 0.2$ or PSNR $\leq 20$, we discard $P$ and move to the next sliding window with stride $128$. 
If there are no ``failed'' crops, $P$ slides to the next non-overlapping position.
In the end, we also manually filter out the crops that are misaligned on and near the fences by visual comparison between input and aligned ground-truth. 
We also manually filter out crops consisting of mostly homogeneous regions (sky, land, sand), to promote diversity in our dataset. 
Our final real burst dataset consists of $185$ \res{} input bursts with corresponding ground truth key-frames from $29$ scenes. Samples from our real burst dataset are shown in Figure~\ref{fig:real_bursts}.

\section{Task Specificity and Comparison with SOLD~\cite{liu2021learning}}
\begin{figure}[ht]
    \centering
    \def\wf{0.32}
    
    \includegraphics[width=\wf\linewidth]{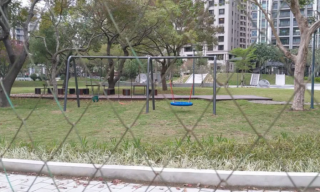}
    \includegraphics[width=\wf\linewidth]{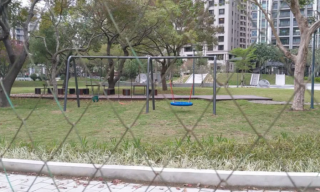}
    \includegraphics[width=\wf\linewidth]{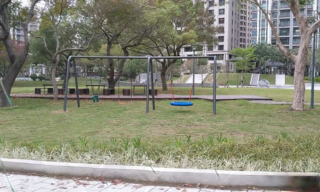}

    \includegraphics[width=\wf\linewidth]{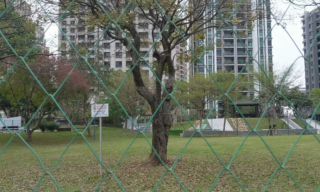}
    \includegraphics[width=\wf\linewidth]{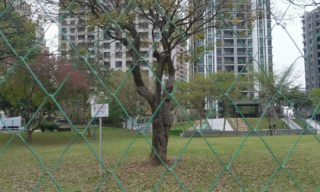}
    \includegraphics[width=\wf\linewidth]{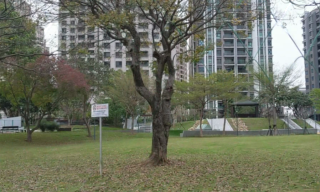}

    \includegraphics[width=\wf\linewidth]{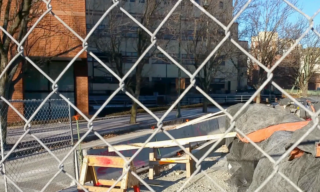}
    \includegraphics[width=\wf\linewidth]{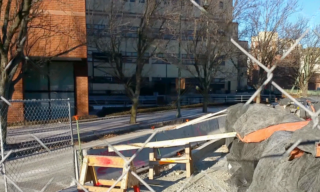}
    \includegraphics[width=\wf\linewidth]{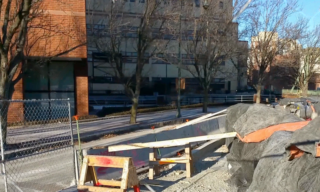}
    \begin{subfigure}{\wf\linewidth}
        \includegraphics[width=\textwidth]{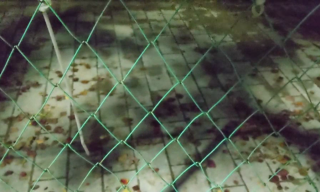}
        \caption{Keyframe}
    \end{subfigure}
    \begin{subfigure}{\wf\linewidth}
        \includegraphics[width=\textwidth]{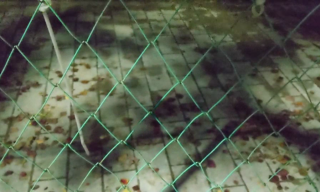}
        \caption{Output (original)}
    \end{subfigure}
    \begin{subfigure}{\wf\linewidth}
        \includegraphics[width=\textwidth]{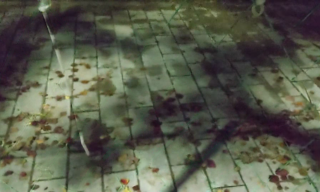}
        \caption{Improved output}
    \end{subfigure}
    % \begin{subfigure}{\wf\linewidth}
    %     \includegraphics[width=\textwidth]{figs/flow_gt_cropped.jpg}
    %     \caption{Output}
    % \end{subfigure}
    \caption{Better data augmentation can make the fence segmentation network more robust to varied types of fences, 
    thus improving the quality of frame inpainting on real sequences \emph{without} the need for online finetuning.}
    \label{fig:newseg}
\end{figure}

One potential criticism towards our approach is our focus on a specific type of 
obstruction (fences), and the fact that we heavily rely on a specific prior (pre-trained fence segmentation model), 
which can harm generalization to new inputs, not commonsly seen in the training data.
In comparison, SOLD~\cite{liu2021learning} is a multi-frame approach can handle various types of obstructions.
% but requires some architectural changes and access to appropriate training datasets to do so.
% In addition, SOLD is also limited when faced with atypical obstructions (e.g., fences), requiring a 
However, SOLD is also limited when faced with atypical obstructions (e.g., fences), requiring
scene-specific, costly online optimization that takes $\sim 3$ minutes, 
to achieve good results, making it impractical for real-world application.
Our method trades-off generality for reconstruction and runtime performance (the latter is a feature missing from previous de-fencing works), 
producing better de-fencing results than SOLD, at a fraction of its runtime, without requiring scene-specific optimization. 
Besides, de-fencing is an important problem in its own right, with an extensive literature in computer vision (see Section~2.1 in the main paper).
% Focusing on this particular type of obstruction allowed us to develop an approach that is both effective 
% \emph{and} efficient (the latter is a feature missing from almost all previous de-fencing works), 
% and can be applied in real world scenaria. 
Finally, we can make our method more robust to a broader variety of fences (e.g., rhombic rotated fences, etc.) by improving our data augmentation protocol.
To showcase this, we have added more scale, rotation, shape and color variation during the training of the fence segmentation model.
As shown in Figure~\ref{fig:newseg}, after adding these additional data augmentations, the fence segmentation model 
can accurately segment fences that are rotated, very thin fences, or have low contrast with respect to 
the background, subsequently improving de-fencing quality.
Extending our method to handle other types of obstructions (e.g., reflections), 
is also a direction we are currently exploring.

\section{Qualitative Results}
\begin{figure*}
    \def\wf{0.16}
    \centering
    \includegraphics[width=\textwidth]{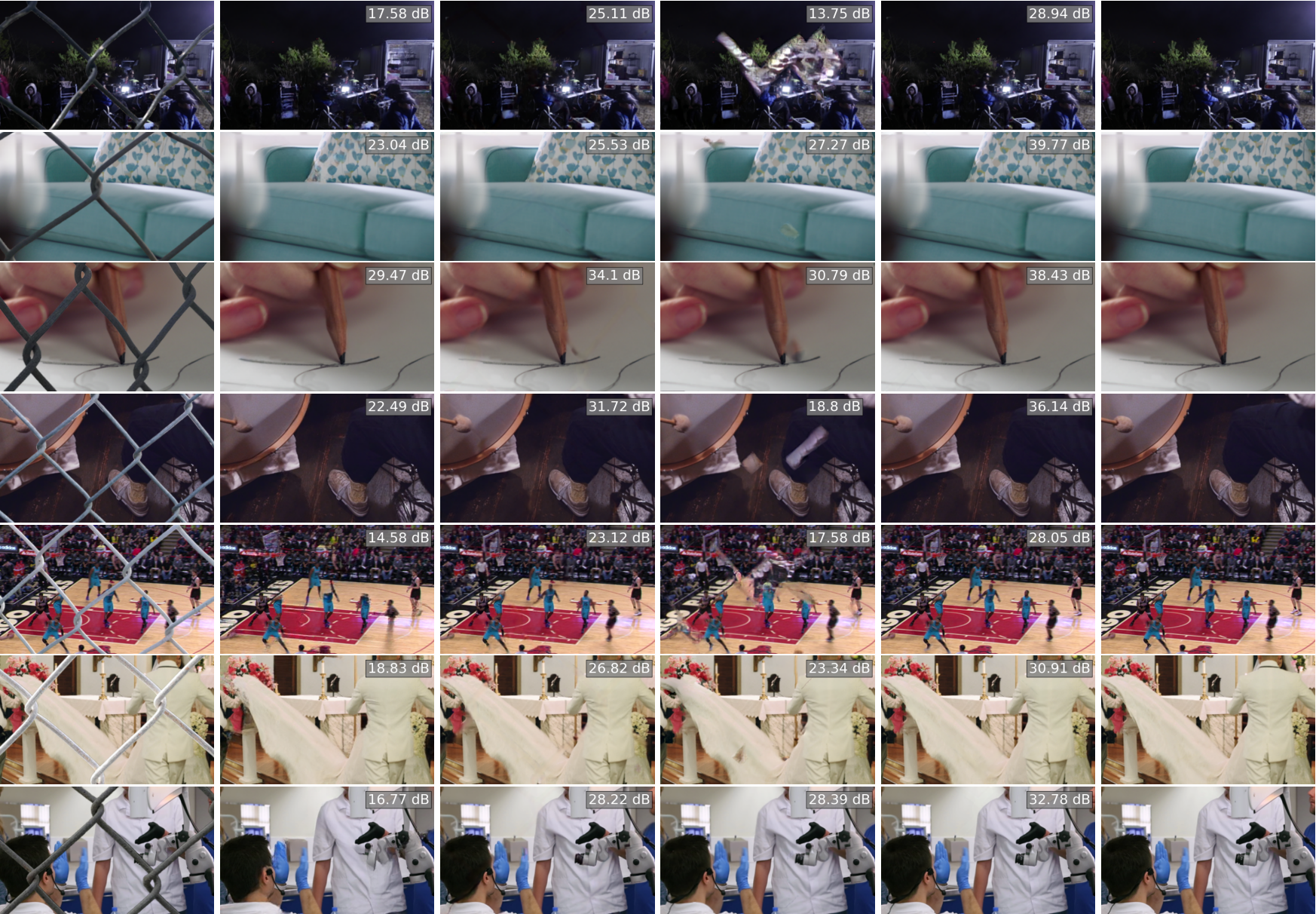}
    \begin{subfigure}{\wf\textwidth}
        \caption{Keyframe}
    \end{subfigure}
    \begin{subfigure}{\wf\textwidth}
        \caption{LaMa~\cite{suvorov2022resolution}}
    \end{subfigure}
    \begin{subfigure}{\wf\textwidth}
        \caption{SOLD~\cite{liu2021learning}}
    \end{subfigure}
    \begin{subfigure}{\wf\textwidth}
        \caption{FGVC~\cite{gao2020flow}}
    \end{subfigure}
    \begin{subfigure}{\wf\textwidth}
        \caption{Ours}
    \end{subfigure}
    \begin{subfigure}{\wf\textwidth}
        \caption{Ground truth}
    \end{subfigure}
    \caption{De-fencing results on sequences from our \textbf{synthetic} data, 
        and respective PSNR scores \emph{inside} the fence mask area. 
        The leftmost column shows the obstructed keyframe, and the next 5 rows show its reconstructed
        version using various baselines and our approach. Zoom in to notice differences in reconstructed frames. 
    }    
    \label{fig:more_qualitative_results_synthetic}
\end{figure*}

\begin{figure*}
    \centering
    \includegraphics[width=\textwidth]{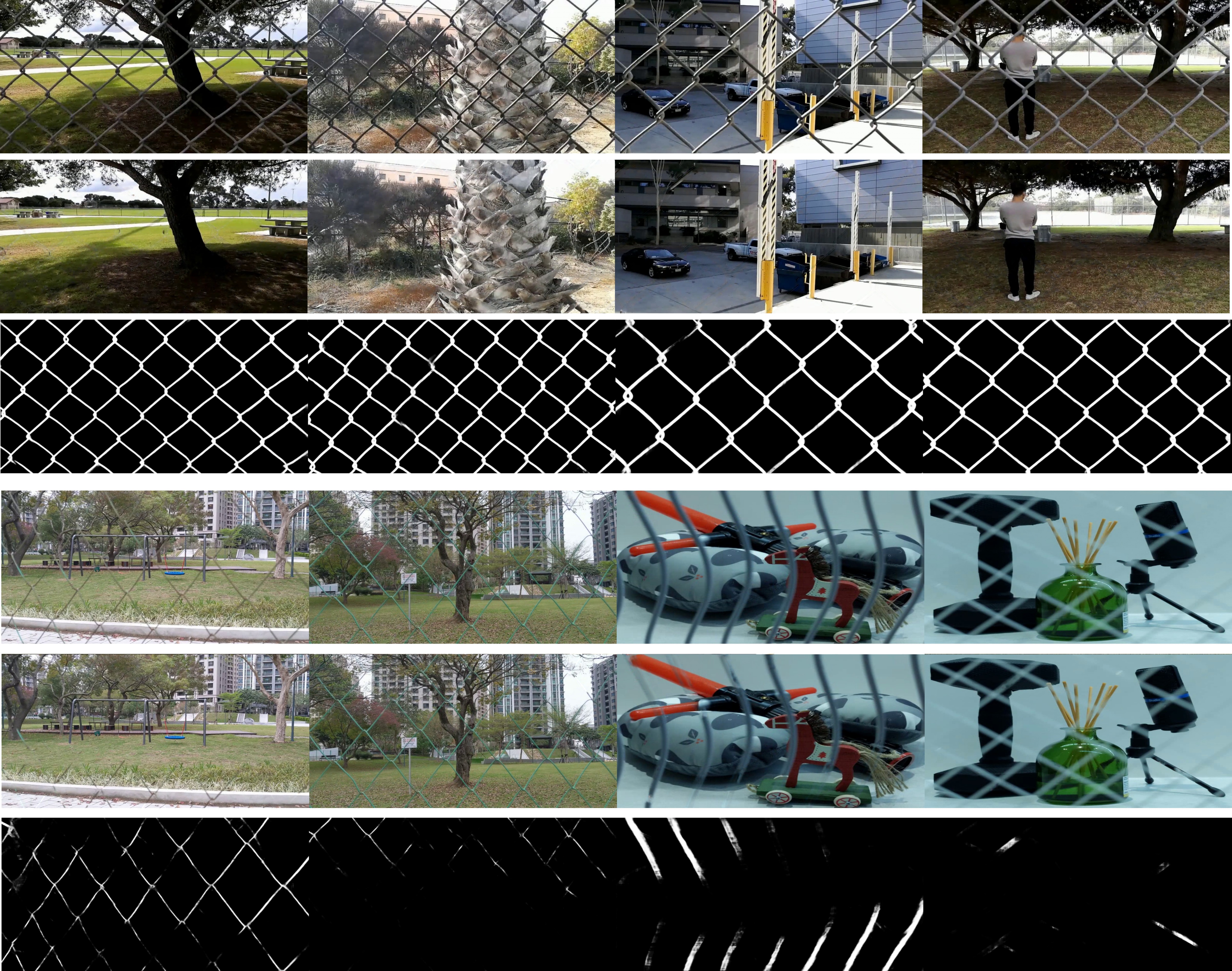}
    \caption{De-fencing results on sequences from~\cite{xue2015computational,liu2021learning}.
        From top to bottom: obstructed keyframe, reconstructed keyframe using our approach, 
        estimated fence segmentation using our U-net fence segmentation model.
        The second group of results shows failure cases: when the fence obstruction is outside our training distribution 
        (e.g., scale - very thin fences, irregular fence pattern, such as vertical bars, extreme blur etc.)
        the fence segmentation estimation fails, affecting reconstruction quality. 
        Addressing unusual fence obstructions 
        like these is our main focus for future work.
    }   
    \label{fig:more_qualitative_results_real1}
\end{figure*}

\begin{figure*}
    \def\wf{0.16}
    \centering

    \includegraphics[width=\textwidth]{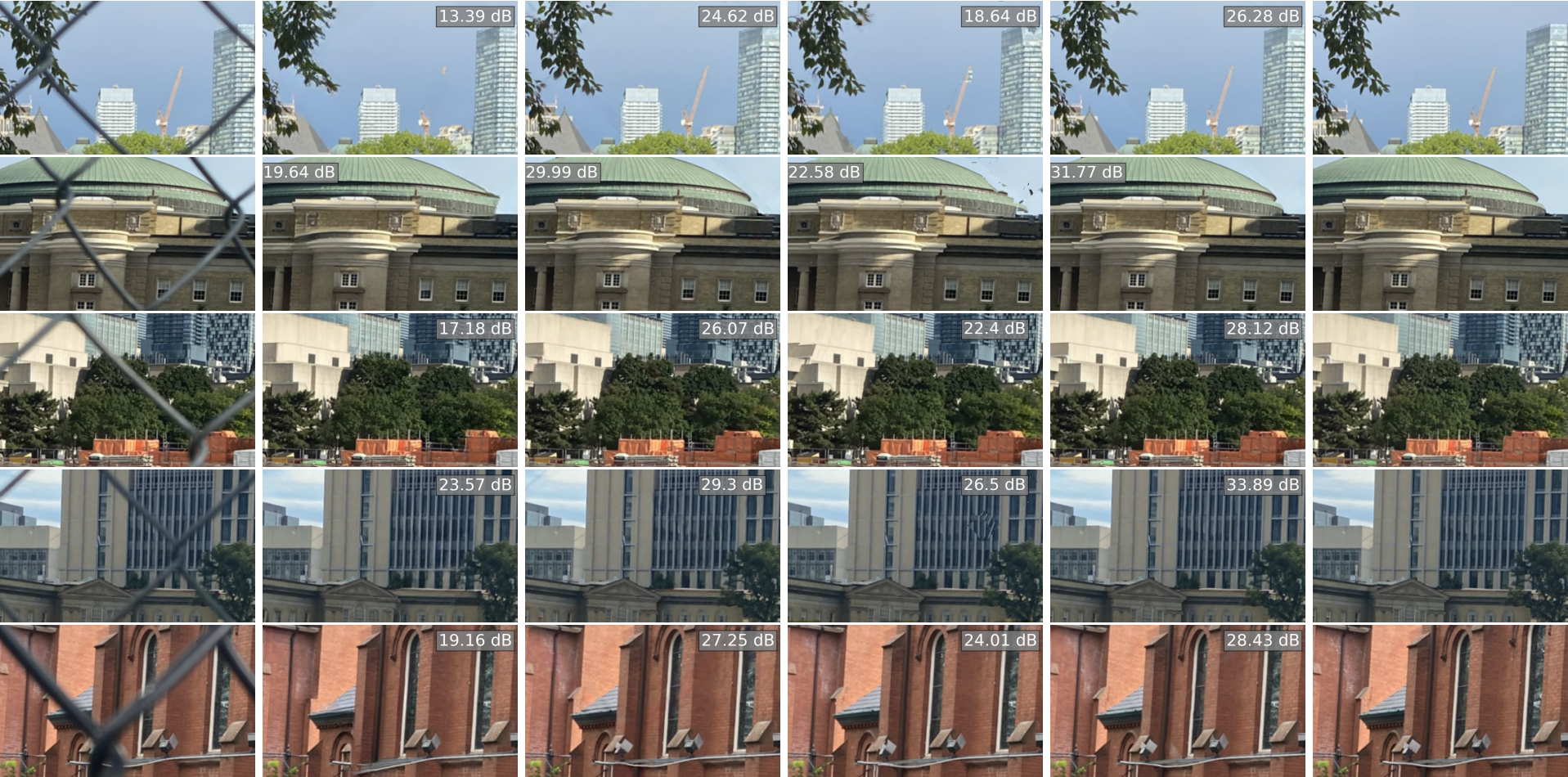}
    \begin{subfigure}{\wf\textwidth}
        \caption{Keyframe}
    \end{subfigure}
    \begin{subfigure}{\wf\textwidth}
        \caption{LaMa~\cite{suvorov2022resolution}}
    \end{subfigure}
    \begin{subfigure}{\wf\textwidth}
        \caption{SOLD~\cite{liu2021learning}}
    \end{subfigure}
    \begin{subfigure}{\wf\textwidth}
        \caption{FGVC~\cite{gao2020flow}}
    \end{subfigure}
    \begin{subfigure}{\wf\textwidth}
        \caption{Ours}
    \end{subfigure}
    \begin{subfigure}{\wf\textwidth}
        \caption{Ground truth}
    \end{subfigure}
    \caption{Qualitative de-fencing results on \textbf{real} sequences, 
        and respective PSNR scores \emph{inside} the fence mask area. 
        The leftmost column shows the obstructed keyframe, and the next 5 rows show its reconstructed
        version using various baselines and our approach. Zoom in to notice differences in reconstructed frames.
    }    
    \label{fig:more_qualitative_results_real2}
\end{figure*}

Figure~\ref{fig:more_qualitative_results_synthetic} shows additional qualitative results on sequences from our synthetically generated test set.
Figure~\ref{fig:more_qualitative_results_real1} shows results on real sequences, taken from previous works~\cite{xue2015computational,liu2021learning}.
We are also including some failure cases, where the fence segmentation model encounters fences at scales or shapes that are out of its training distribution, 
resulting in low de-fencing quality.
Finally, in Figure~\ref{fig:more_qualitative_results_real2} we compare results from our method and other baselines on examples from the real burst dataset
we collected.

\end{document}